\pdfoutput=1

\documentclass[11pt]{article}
\usepackage[utf8]{inputenc}
\usepackage{CJKutf8}
\usepackage{listings}
\usepackage{xcolor}
\usepackage{booktabs}
\usepackage{multirow}
\usepackage{multicol}
\usepackage{array}
\usepackage{amssymb} 
\usepackage{xcolor} 
\usepackage{siunitx} 
\usepackage{tabularray}
\usepackage{EMNLP2023}
\usepackage{amsmath}
\usepackage{times}
\usepackage{latexsym}
\usepackage{graphicx}
\usepackage{longtable} 
\usepackage{ragged2e} 
\usepackage[T1]{fontenc}
\usepackage{enumitem}   %
\usepackage{float}
\usepackage{hyperref}
\usepackage{url}
\usepackage{array}        
\usepackage[table]{xcolor}       
\usepackage{multirow}     
\usepackage{graphicx}     
\usepackage{booktabs}     
\usepackage[skins]{tcolorbox}
\usepackage{geometry}
\usepackage{setspace}
\usepackage{fancyvrb}
\usepackage{minted}
\usepackage{listings}
\setlist[itemize]{nosep, before=\vspace{-0.5em}, after=\vspace{-0.5em}}
\setlist[enumerate]{nosep, before=\vspace{-0.5em}, after=\vspace{-0.5em}}
\usepackage[utf8]{inputenc}
\usepackage{algorithm} 
\usepackage{algorithmicx} 
\usepackage{algpseudocode} 
\usepackage{microtype}

\usepackage{inconsolata}
\usepackage[table]{xcolor} 
\usepackage{arydshln} 

\definecolor{lightblue}{rgb}{0.9, 0.95, 1} 

\title{MotivGraph-SoIQ: Integrating Motivational Knowledge Graphs and Socratic Dialogue for Enhanced LLM Ideation}

\usepackage{amsmath}

\author{
 \textbf{Xinping Lei\textsuperscript{1,4,*}},
 \textbf{Tong Zhou\textsuperscript{1,*}},
 \textbf{Yubo Chen\textsuperscript{1,2,3,\dag}},
 \textbf{Kang Liu\textsuperscript{1,2}},
 \textbf{Jun Zhao\textsuperscript{1,2}}
    \\
    \textsuperscript{1} The Key Laboratory of Cognition and Decision Intelligence for Complex Systems \\
    \textsuperscript{2} School of Artificial Intelligence, University of Chinese Academy of Sciences, Beijing, China \\
    \textsuperscript{3} Hunan Provincial Key Laboratory of Philosophy and \\ Social Sciences of Artificial Intelligence and Precision International, Hunan Normal University \\
    \textsuperscript{4} Beijing University of Posts and Telecommunications\\
    \small{
   lxp@bupt.edu.cn, tongzhou21@outlook.com, \{yubo.chen, kliu, jzhao\}@nlpr.ia.ac.cn
 }
}

\begin{document}
\maketitle
\def\thefootnote{*}\footnotetext{These authors contribute equally to this work.}\def\thefootnote{\arabic{footnote}}
\def\thefootnote{\dag}\footnotetext{Corresponding authors.}\def\thefootnote{\arabic{footnote}}
\begin{abstract}
Large Language Models (LLMs) hold substantial potential for accelerating academic ideation but face critical challenges in grounding ideas and mitigating confirmation bias for further refinement. We propose integrating motivational knowledge graphs and socratic dialogue to address these limitations in enhanced LLM ideation (MotivGraph-SoIQ). This novel framework provides essential grounding and practical idea improvement steps for LLM ideation by integrating a Motivational Knowledge Graph (MotivGraph) with a Q-Driven Socratic Ideator. The MotivGraph structurally stores three key node types—problem, challenge, and solution—to offer motivation grounding for the LLM ideation process. The Ideator is a dual-agent system utilizing Socratic questioning, which facilitates a rigorous refinement process that mitigates confirmation bias and improves idea quality across novelty, experimental rigor, and motivational rationality dimensions. On the ICLR25 paper topics dataset, MotivGraph-SoIQ exhibits clear advantages over existing state-of-the-art approaches across LLM-based scoring, ELO ranking, and human evaluation metrics.
\end{abstract}

\section{Introduction}

The potential of Large Language Models (LLMs) in supporting academic research \cite{achiam2023gpt,yang2024qwen2,liu2024deepseek, Chen_2024} within the domain of scholarly research has garnered increasing attention \cite{radensky2024scideator,si2024can,lu2024ai,gupta2025all}. This includes the automated generation of literature reviews \cite{liang2025surveyx,azaria2023chatgptremarkabletool}, assistance in experimental design, and the enhancement of academic writing \cite{lu2024ai,weng2024cycleresearcher}. Notably, leveraging the creativity of large language models to generate novel research ideas \cite{wang2024scipip,li2024chain,si2024can,baek2024researchagent} is particularly compelling, which promises to accelerate the process of knowledge discovery, aiding researchers in transcending conventional thinking patterns and expanding the frontiers of exploration \cite{gottweis2025towards}. However, the practical application of LLMs for generating research ideas still confronts two critical bottlenecks. Firstly, the generation process lacks a robust theoretical or factual grounding, which makes it challenging to create innovative and feasible ideas. Secondly, the issue of confirmation bias makes it difficult for LLMs to improve ideas.
        
\paragraph{Motivation Grounding}
Human researchers establish academic motivation connections through an extensive literature review, which helps uncover their underlying motivations and problem-solving approaches. This process enables them to navigate complex knowledge domains, understand fundamental concepts, and promote innovation across different disciplines. For example, researchers may observe that ant colonies utilize pheromone trails to identify optimal paths to food sources. Simultaneously, they recognize the challenge of optimizing data routing in large-scale wireless sensor networks. By linking these insights, they form an academic motivation connection between biological swarm intelligence and network optimization. Such a connection can lead to the novel application of ant colony optimization algorithms to improve routing efficiency in sensor networks. The effectiveness of academic motivation connections lies in their ability to foster a comprehensive understanding of disparate fields and encourage combinatorial innovation, thereby generating novel and valuable ideas.

However, the internal knowledge of Large Language Models is probabilistic \cite{ye2025open}, unstable \cite{atil2024llm}, and inherently biased due to training data distribution. Relying solely on large language models to generate ``academic motivation connections'' can lead to unreliable innovation. Concerns persist that LLM-generated ideas may be primarily hallucinatory, superficial \cite{gupta2025all}, or infeasible \cite{si2024can}. Although approaches have been proposed to ground LLMs with external academic resources for background information \cite{lu2024ai}, their limited context windows hinder effective processing of extensive literature and the formation of deep connections. Consequently, enabling LLMs to generate innovative and feasible ideas necessitates a motivation knowledge base capable of providing a profound grasp of academic research's underlying motivations and relationships, in a format compatible with LLM processing characteristics.

\paragraph{Confirmation bias}
Confirmation bias \cite{nickerson1998confirmation} is a cognitive bias where individuals favor information that confirms pre-existing beliefs \cite{wason1968reasoning}. Human researchers are susceptible to favoring data that supports their hypotheses, sometimes overlooking contradictory evidence. Discussions between the mentor and the researcher are crucial for its mitigation in scientific contexts. In these settings, researchers present their hypotheses and reasoning to their mentors, who challenge assumptions, question methodologies, and highlight overlooked counterexamples, helping to correct biased reasoning and flawed assumptions. LLMs also exhibit this bias, struggling with novel thought generation and self-correction once an initial stance is established \cite{liang2023encouraging,zhao2024awecita}. A key challenge in leveraging LLMs for academic ideation is enabling them to identify critical weaknesses in their generated ideas. While effective for superficial issues, current self-reflection methods fail to address fundamental shortcomings such as incorrect assumptions due to their vulnerability to confirmation bias \cite{liang2023encouraging}. Thus, developing strategies for LLMs to refine ideas while actively mitigating this bias remains a considerable challenge.

In this paper, we propose Socratic LLM Ideation with Academic Motivation Graph (MotivGraph-SoIQ) to address the challenges above.

We propose the Motivational Knowledge Graph (MotivGraph) as a foundation for grounded idea generation. To build this graph, we develop Science Motivation Miner, which automatically extracts (problem, challenge, method) triplets from published papers and organizes them into interconnected nodes and edges. During ideation, our autonomous multi-tool framework guides LLMs to query and update the MotivGraph at each step, ensuring that generated ideas reference specific graph nodes and include explicit source annotations. By grounding every concept in the underlying literature, this approach enhances traceability and strengthens the validity of the generated ideas. We propose the Q-Driven Socratic Ideator to enhance idea quality further and mitigate confirmation bias. Inspired by the Socratic method \cite{benson2011socratic,leigh2007platonic}, an LLM acts as a "mentor" to critically question a "researcher" agent. The mentor assesses logic, self-consistency, and rigor, while the researcher leverages structured domain knowledge to generate ideas. This dialogue prompts the researcher to rectify flaws, thereby avoiding confirmation bias inherent in self-reflection and reducing extra external knowledge requirements, simplifying the refinement process. We evaluate MotivGraph-SoIQ on a topic set constructed from ICLR25 papers and compare it against a strong baseline by having DeepSeek-V3 generate ideas under both approaches. Using DeepSeek-V3–generated proposals, our method achieves 10.2 \% higher novelty and 6 \% higher motivational rationality in LLM-based scoring, yielding an average LLM ELO score that is 38 points above the baseline. Human evaluations of the same DeepSeek-V3–generated ideas confirm these gains, showing increases of 7.98 \% in novelty and 5.56 \% in motivational rationality. Across all metrics, MotivGraph-SoIQ consistently outperforms the baseline.

Our main contributions are summarized as follows:
\begin{enumerate}[ 
    label=\textbf{\arabic*:},     
    left=0pt,                                   
    labelsep=0.5em,                               
    itemsep=0.75\baselineskip                   
]
    \item To address the lack of motivational grounding and limited self-improvement in LLM-based ideation, we propose \textbf{MotivGraph-SoIQ}. This unified framework integrates a Motivational Knowledge Graph with a Socratic ideation loop to produce grounded, high-quality ideas.
    \item We introduce \textbf{SciMotivMiner} to tackle the challenge of constructing a structured motivational resource from literature. SciMotivMiner automatically extracts (problem, challenge, method) triplets from published papers to build the MotivGraph, enabling motivational grounding for idea generation.
    \item We develop the \textbf{Q-Driven Socratic Ideator} to handle the difficulty of refining ideas and mitigating biases. This module employs a questioning-based self-improvement loop with four specialized tools for compelling graph exploration and strategic novelty injection, improving idea quality across multiple evaluation metrics.
    \item We conduct concise experiments on a topic set from ICLR25 papers, demonstrating that MotivGraph-SoIQ significantly outperforms strong baselines in novelty, experimental feasibility, motivational rationality, and diversity, achieving a 10.2 \% improvement in novelty, a 6 \% improvement in motivation, and an average ELO score 38 points higher.
\end{enumerate}

\section{Method}
In this section, we detail our LLM-based ideation methodology, the MotivGraph-SoIQ Framework, which integrates two core components:
(i) \textbf{MotivGraph}, a motivation-enhancing knowledge graph for structured motivation representation, and
(ii) \textbf{Q-Driven Socratic Ideator}, an adversarial agentic system that refines ideas through ``Socratic questioning'' and ``maieutics''.

\subsection{MotivGraph}

The \textbf{MotivGraph} serves two primary purposes. Firstly, it provides the underlying knowledge base to supply relevant knowledge crucial for the ideation process. Secondly, the explicit relationships between entities within the graph offer concrete examples of how problems can be framed and addressed. This structure is a valuable source of inspiration, specifically aiding LLMs in formulating clear and compelling motivations for novel research ideas.

\subsubsection{MotivGraph construction}
\label{MotivGraph construction}
Amabile's Componential Theory of Creativity \cite{amabile1996creativity}posits that motivation constitutes one of the three essential components of innovation (alongside domain-relevant skills and creativity-relevant processes), with intrinsic motivation being particularly critical for breakthrough ideation. We design the \textbf{MotivGraph} as a graph structure consisting of three principal node types: \textit{problem}, \textit{challenge}, and \textit{solution}. A \textit{problem} node signifies a minimally granular research topic or task, a \textit{challenge} node indicates a specific difficulty encountered within a \textit{problem}, and a \textit{solution} node represents a concrete method addressing a \textit{challenge}. \textbf{Motivation} information is represented by triples formed through inter-node connections, specifically in the format (problem, challenge, solution). Each node is further characterised by two attributes: a concise and precise name for unique identification, and a description that provides further detail and aids in the graph's semantic representation, matching, and retrieval processes.

\begin{figure}
    \centering
    \includegraphics[width=1\linewidth]{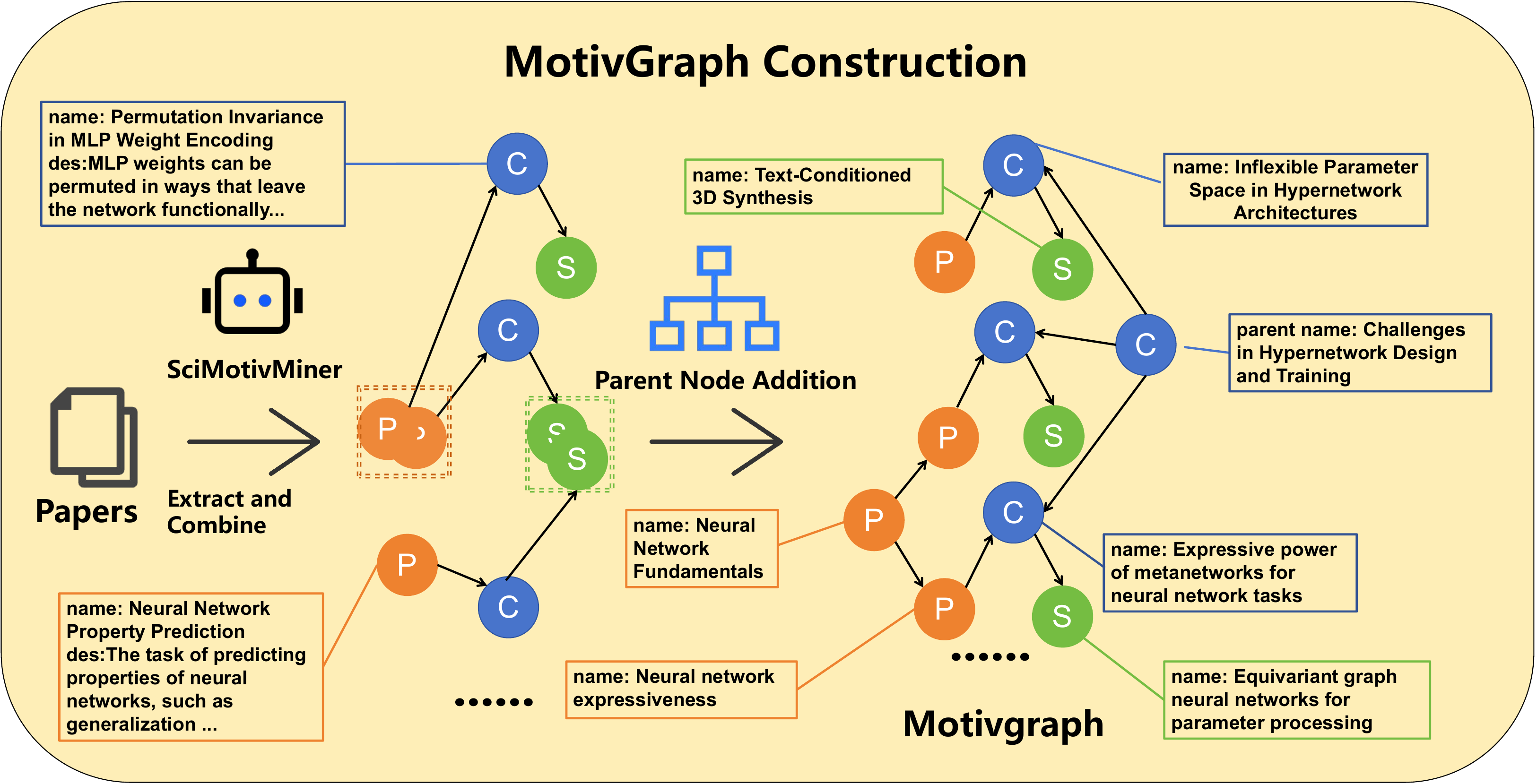}
    \caption{motivgraph construction figure}
    \label{motivgraph figure}
\end{figure}

The MotivGraph is represented as a graph $G = (V, E)$, where $V$ is the set of nodes and $E$ is the set of edges. The nodes $V$ are classified into three types: problem ($P$), challenge ($C$), and solution ($S$). The edge set $E$ includes three distinct edge types: \textit{parent-of} (for hierarchical links), \textit{problem-challenge} (connecting $P$ to $C$), and \textit{challenge-solution} (connecting $C$ to $S$). Figure \ref{motivgraph figure} shows the construction process of MotivGraph.The specific construction method will be introduced in the following sections.

\subsubsection{SciMotivMiner}

For each scientific paper $P$, we employ our method, SciMotivMiner, denoted as $\text{SMM}(P)$, to process the paper and identify triples of related problems, challenges, and solutions. Consequently, $\text{SMM}(P)$ outputs a set of $n$ distinct (Problem, Challenge, Method) triples: $\{(P_i,C_i, S_i)\}_{i=1}^n = \text{SMM}(P)$.

For each extracted  $(P_i, C_i, S_i)$, SciMotivMiner summarizes a concise entity name and a brief description. The naming process adheres to the rules to ensure clarity and consistency within the knowledge graph. The exact rules are detailed in the appendix \ref{SciMotivMiner rules}.

These stringent rules ensure a standardized, informative, and author-name-agnostic representation of research motivation and proposed solutions within the SciMotivMiner knowledge graph. For identical nodes, SciMotivMiner will merge them.

\paragraph{Hierarchical Parent Node Addition}
\label{Hierarchical Parent Node Addition}
Academic problems and challenges inherently possess a hierarchical structure, with different papers addressing varying granularities. To capture these relationships and prevent knowledge fragmentation within the MotivGraph, we introduce Hierarchical Parent Node Addition for both Problem (P) and Challenge (C) entities. This process organizes knowledge into a coherent hierarchy, crucial for practical exploration.

Our Parent Node Addition Algorithm operates iteratively. It begins by embedding all initial Problem/Challenge nodes into a vector space. The algorithm then repeatedly selects a focal node, identifies its k most similar neighbors within the current working set, and employs an LLM to evaluate their semantic coherence for merging. If the LLM deems an add appropriate, a new, more general parent node is created and linked to its children by parent-of edges. Processed nodes are then removed from the working set. This dynamic process ensures each node is considered for forming a parent at most once, building a multi-level abstract representation of the concepts. See the appendix \ref{Hierarchical Parent Node Addition} for details.

\subsection{Q-Driven Socratic Ideator}
The \textbf{Q-Driven Socratic Ideator} is a dual-agent system consisting of a mentor agent and a researcher agent. Its operational principles are inspired by Socratic questioning and maieutics. The mentor agent adheres to the elenchus (Socratic refutation) through triple-axis questioning---probing innovation, feasibility, and rationality---thereby exposing logical gaps without prescriptive solutions. The researcher agent operationalizes maieutics (intellectual midwifery) by synthesizing knowledge through: (1) introspective retrieval of dialogue history (``knowledge amniotic fluid''), and (2) external tool-augmented searches, ultimately ‘giving birth’ to refined ideas through self-directed epistemic labor. The following subsections detail the architecture, roles, and interaction dynamics of the agents within this system. The following sections delineate the two-phase architecture of the Q-Driven Socratic Ideator: (i) the \textbf{Exploration Phase}, and (ii) the \textbf{Deliberation Phase}.

\begin{figure*}[htbp]
    \centering
    \includegraphics[width=1.0\linewidth]{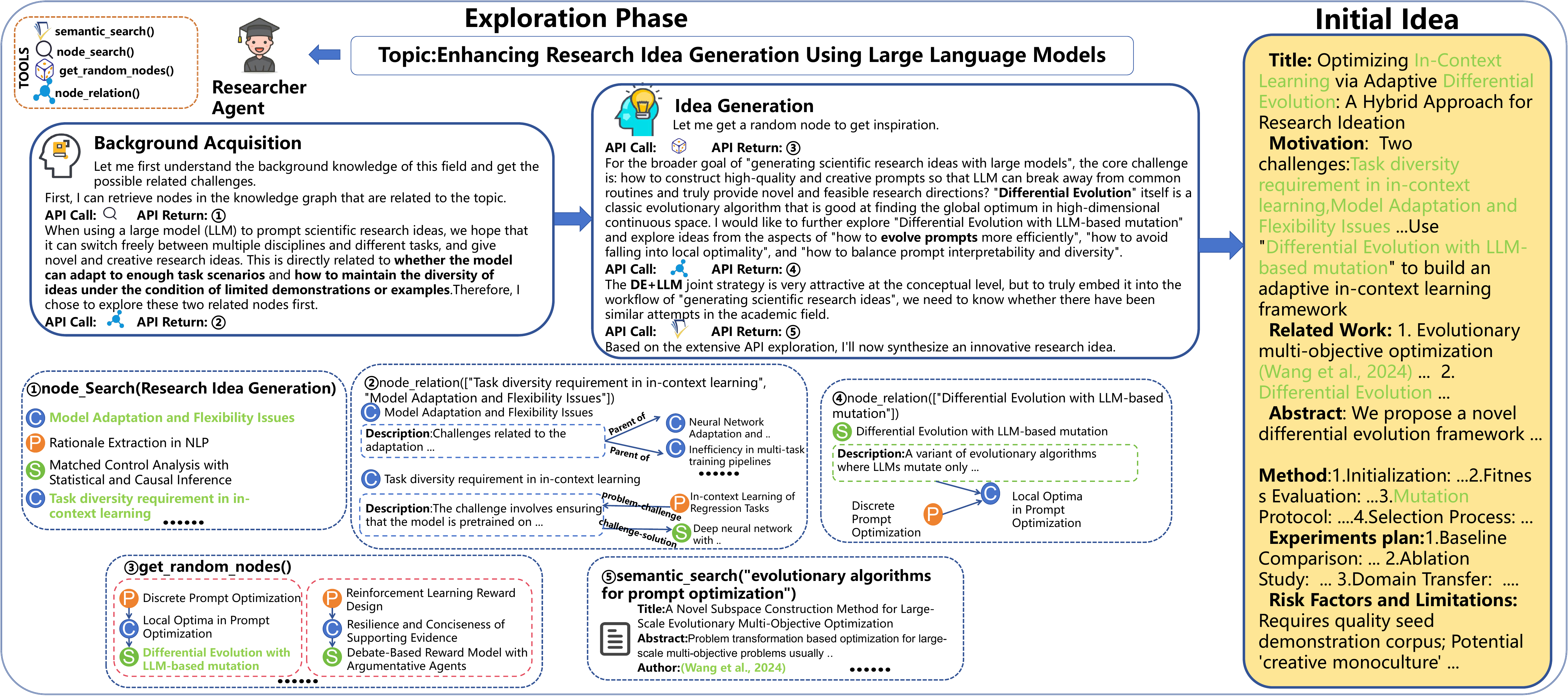}
    \caption{Exploration Phase Pipeline}
    \label{Exploration Phase Pipeline}
\end{figure*}

\subsubsection{Exploration Phase}
The researcher agent primarily carries out the Exploration Phase. Based on the provided target domain or task description, the researcher agent performs knowledge exploration and generates innovative ideas. Figure \ref{Exploration Phase Pipeline} shows the process of the Exploration Phase. See Appendix \ref{Detailed Researcher Agent's Toolset} for further details.

\paragraph{Knowledge Exploration and Ideation}
We designed three API tools to help the researcher agent better understand the target domain or task and generate an idea:

\paragraph{Graph Node Fuzzy Search} This tool allows the researcher agent to obtain an overall understanding of the target domain/task by fuzzy-matching and retrieving related \textit{problem}, \textit{challenge}, and \textit{solution} entities from the MotivGraph based on a search query.

\paragraph{Graph Node Relation Retrieval} By providing an interesting node's name, the researcher agent can retrieve its description and neighboring nodes, gaining hierarchical relationships and (problem, challenge, solution) motivation triplets. This deepens understanding and supports effective subsequent retrieval.

\paragraph{Semantic Scholar Literature Search} This API provides query-based literature search, offering more specific information than the graph for a comprehensive understanding of particular challenges or technologies.

\paragraph{Get Random Nodes to Enhance Novelty} After sufficient knowledge exploration, the researcher agent uses this API to obtain random problem-challenge-solution triples. It then attempts to apply these to the target domain, seeking potential connections or adaptations. This mechanism supports the "creativity-relevant processes" from Amabile's Componential Theory of Creativity \cite{amabile1996creativity}, ensuring idea novelty. Simultaneously, the inherent logic of the MotivGraph's (problem, challenge, solution) triples fosters "intrinsic motivation," driving the agent to explore adaptations of external nodes to the target domain, facilitating the discovery of new problems or innovative solutions.

\subsubsection{Deliberation Phase}

\begin{figure}
    \centering
    \includegraphics[width=1\linewidth]{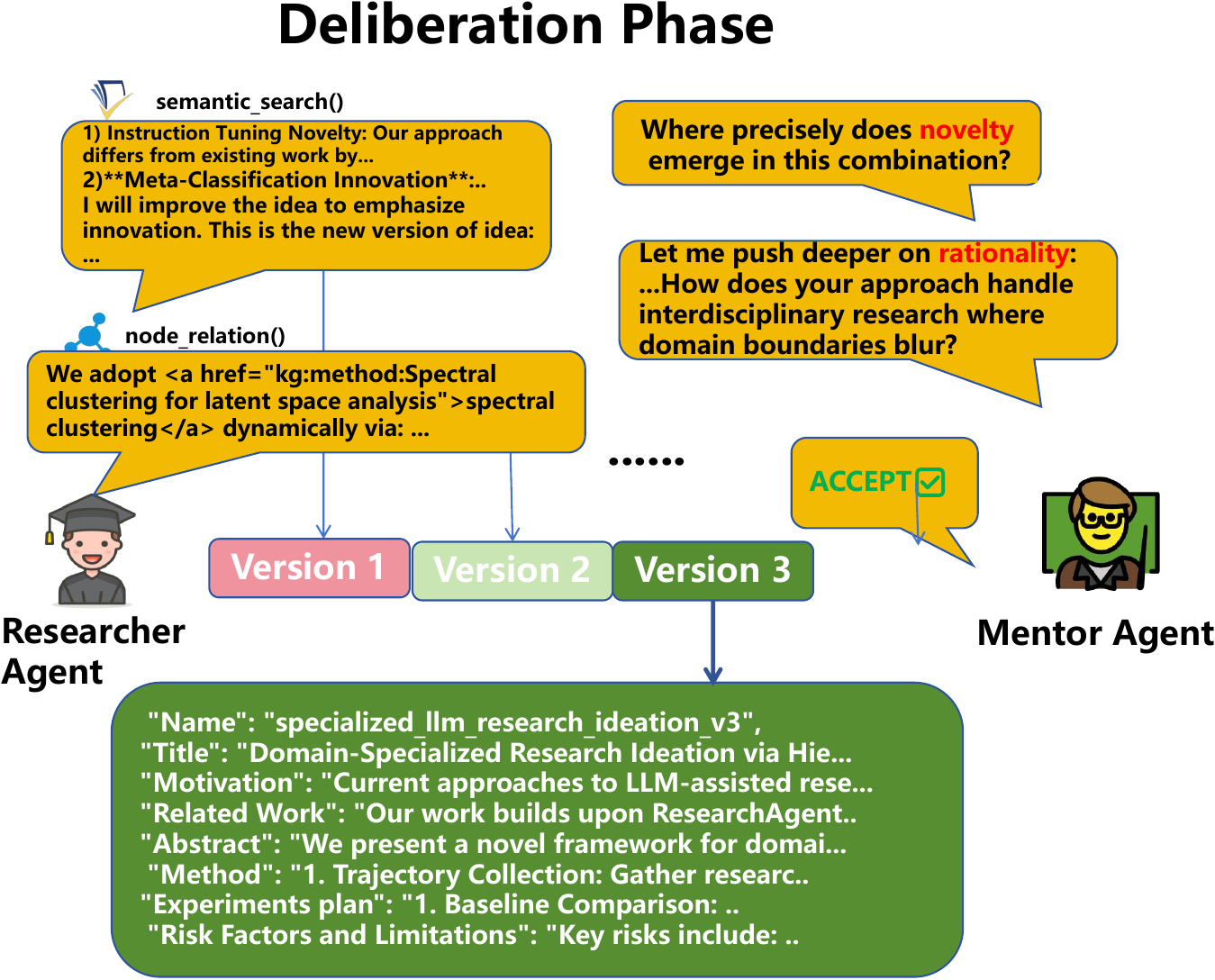}
    \caption{Deliberation Phase Pipeline}
    \label{Deliberation Phase Pipeline}
\end{figure}

Following the initial Exploration Phase, the researcher agent enters the Deliberation Phase, engaging in multi-round deliberation with the mentor agent. This phase is designed to rigorously evaluate and refine previously generated innovative ideas, embodying the core principles of Socratic interaction. Figure \ref{Deliberation Phase Pipeline} shows the process of the deliberation phase.

During this phase, the mentor agent challenges the researcher agent's idea from three predefined angles, acting as a form of Socratic elenchus (refutation). These angles are: \textit{innovation}, \textit{feasibility}, and \textit{rationality}. The mentor agent poses probing questions to test the idea's robustness and underlying assumptions, such as: ' How does this solution transcend prior ideas?’ (\textit{Innovation})‘What tools would implement this?’ (\textit{Feasibility})‘Why is your method effective?’(\textit{Rationality})

The researcher agent responds by providing justification and defending its idea, drawing upon the knowledge accumulated during the Exploration Phase and the reasoning process that led to the idea's generation. In defending its idea, the researcher agent may identify flaws or gaps in its concept or understanding through critical self-reflection. When this occurs, the researcher agent can perform supplementary knowledge exploration (utilizing the available API tools) to seek methods for addressing these weaknesses and gather supporting evidence. Subsequently, the researcher agent presents the refined idea and its updated rationale, awaiting further questioning from the mentor agent. This guided self-correction and refinement process represents the maieutic process, where the agent is guided towards a more robust concept.

Crucially, the mentor agent acts strictly as a critical evaluator and questioner, facilitating the researcher's learning and refinement without providing direct answers or solutions. It does not perform its knowledge gathering or directly modify the researcher agent's ideas. Its role is limited to posing questions based on the predefined angles and the perspective inherent in its role to guide the researcher agent's refinement process.

We can set a specific number of deliberation rounds in advance, and the mentor agent has the flexibility to end the process early if particular criteria are met, like when an idea proves to be strong enough or is not viable after multiple discussions. After the deliberation (either by reaching the round limit or early termination), the mentor agent provides a final overall evaluation: \texttt{ACCEPT} or \texttt{REJECT}. Rejected ideas are discarded. This ensures that the system avoids retaining ideas that cannot be sufficiently justified or improved during deliberation, particularly when the initial random input was challenging to integrate effectively, upholding a quality standard for accepted ideas.

\paragraph*{Process Formalism}
The iterative deliberation process and outcome can be formally represented. Let $Idea_k$ be the state of the idea after round $k$ ($Idea_0$ is the initially generated idea), and $I_k$ denote the interaction (mentor's question and researcher's response) at round $k$. The idea evolves based on the ideator agent's function $f_{\text{Researcher}}$:
$$Idea_k = f_{\text{Researcher}}(Idea_{k-1}, I_k, \text{Exploration})$$
The deliberation phase concludes at round $N_{final}$ (the maximum predefined rounds or an earlier termination round). The final evaluation, $Eval$, is given by the mentor agent's function $e_{\text{Mentor}}$ based on the final idea state and potentially the dialogue history:
$$Eval = e_{\text{Mentor}}(Idea_{N_{final}}, \text{Dialogue History})$$
where $Eval \in \{\texttt{ACCEPT}, \texttt{REJECT}\}$.

\section{Experiment}

To validate MotivGraph-SoIQ's effectiveness, we conducted both comparative and ablation experiments. We constructed the MotivGraph and an evaluation dataset using publicly available literature. The MotivGraph provides motivational grounding, while the evaluation dataset, comprising 100 diverse ICLR 2025 paper topics and their core ideas, serves as ground truth for assessing generated ideas. Our comparisons against baselines demonstrate MotivGraph-SoIQ's superiority, and ablation studies confirm the effectiveness of individual system components. See the Appendix \ref{Dataset Construction and Evaluation Details} for details on the dataset.

\setlength{\tabcolsep}{3pt} 
\renewcommand{\arraystretch}{1.0} 
\begin{table*}[htbp]
    \centering %
    \normalsize
    {
    \fontsize{11}{9}\selectfont 
    \begin{tabular}{ll *7c c}
        \toprule
        \multirow{2}{*}{Baseline} & \multirow{2}{*}{Model} & \multirow{2}{*}{Diversity} & \multicolumn{3}{c}{LLM-evaluator} & \multicolumn{3}{c}{Human-evaluator} & \multirow{2}{*}{Length} \\
        \cmidrule(lr){4-6} \cmidrule(lr){7-9}  
        & & & Nov. & Exp. & Moti. & Nov. & Exp.     & Moti. & \\
        \midrule
        \multirow{3}{*}{AI-S-v2} & DeepSeek-V3 & 0.27 & 7.22 & 8.07 & 8.21 & 6.35 & 6.25 & 5.85 & 2635 \\
        & Deepseek-R1 & \textbf{0.52} & 7.59 & \textbf{8.36} & 8.30 & / & / & / & 3013 \\
        & Qwen2.5-7B & 0.24 & 6.10 & 7.22 & 7.28 & / & / & / & 3060 \\
        \midrule
        \multirow{3}{*}{AI-Researcher} & DeepSeek-V3 & 0.38 & 7.58 & 7.06 & 7.86 & 6.40 & 6.65 & 6.45 & 4985 \\
        & Deepseek-R1 & 0.32 & 7.94 &  7.65  & 8.19 & / & / & / & 4599 \\
        & Qwen2.5-7B & 0.34 & \textbf{7.14} & 5.76 & 7.29 & / & / & / & 5465 \\
        \midrule
        \multirow{3}{*}{SciPIP} & DeepSeek-V3 & 0.42 & 7.61 & 7.23 & 7.61 & 6.20 & / & 5.05 & 4252 \\
        & Deepseek-R1 & 0.41 & 8.07 & 7.71 & 7.85 & / & / & / & 4230 \\
        & Qwen2.5-7B & 0.35 & 6.51 & 6.04 & 6.46 & / & / & / & 5088 \\
        \midrule
        {CycleResearcher} & {CycleResearcher-12B} & 0.29 & {6.61} & {7.52} & {7.39} & 5.50 & 6.25 & 5.35 & 7189 \\
        \midrule
        \multirow{3}{*}{ResearchAgent} & DeepSeek-V3 & 0.23 & 7.43 & 8.39 & 8.06 & 6.30 & 6.60 & 5.85 & \textbf{15255} \\
        & Deepseek-R1 & 0.25 & 8.02 & 8.33 & 8.17 & / & / & / & \textbf{10204} \\
        & Qwen2.5-7B & 0.17 & 6.88 & \textbf{7.67} & \textbf{7.60} & / & / & / & \textbf{13975} \\
        \midrule
        \multirow{3}{*}{Ours} & DeepSeek-V3 & \textbf{0.45} & \textbf{8.39} & \textbf{8.64} & \textbf{8.70} & \textbf{6.45} & \textbf{6.70} & \textbf{6.70} & 4908 \\
        & deepseek-r1 & 0.45 & \textbf{8.30} & 8.00 & \textbf{8.33} & / & / & / & 4753 \\
        & qwen2.5-7b & \textbf{0.43} & 6.46 & 6.64 & 6.52 & / & / & / & 3698 \\
        \midrule
        {Real Paper} & DeepSeek-V3 & \textbf{1.00} & 6.97 & 8.16 & 7.81 & \textbf{7.08} & \textbf{7.36} & \textbf{8.05} & 5030 \\
        \bottomrule
    \end{tabular}
    \caption{Evaluation Results: We use Fast-reviewer as LLM-evaluator. We manually evaluate and score ideas generated by DeepSeek-V3 using three dimensions: Novelty, experiment, and motivation. Ideas generated by SciPIP do not have experimental designs, so their experiments are not manually evaluated.}
    \label{main_table}
    } 
\end{table*}

\subsection{Comparative Baselines}

To assess the effectiveness of our MotivGraph-SoIQ, we selected several baseline methods for comparison. The criteria for their selection were based on similarities in generating idea components similar to ours (Motivation, Related Work, Abstract, Method, Experiment Plan, Risk Factors, and Limitations) or employing entity/graph-based information enhancement. Please refer to Appendix \ref{baseline} for detailed information. The selected baselines are below:

\textbf{AI-Researcher}: 
This method, proposed in \cite{si2024can}, uses the author's publicly available code to generate ideas.

\textbf{Cycle Researcher (12B)}: Proposed in \cite{weng2024cycleresearcher}, we use the author's publicly available code to generate idea proposals.

\textbf{AI-Scientist-v2}: This is an improved version of AI-Scientist \cite{yamada2025ai}. We use the author's publicly available code to generate ideas.

\textbf{SciPIP}: 
We use the author's publicly available code to generate ideas.

\textbf{ResearchAgent}: 
We reproduce this method following the methodology described in the author's paper \cite{baek2024researchagent}.

\subsection{Ablation Studies}

We conducted a series of ablation studies to understand better each component's contribution to our method's overall performance and validate the necessity of these design choices. This section systematically removed or modified one or more parts of the MotivGraph and the Critique-Driven Agent System. We tested these variants using the same experimental setup and evaluation metrics as the whole method. By comparing the performance of different variants, we can quantify the effectiveness of each component and reveal the key roles they play in the ideation process.

The following ablation variants were tested:

1. \textbf{W/O Mentor}: The deliberation loop involving the mentor agent was removed in this configuration. The researcher agent generates and revises the idea by themselves.

2. \textbf{W/O Graph}: In this experimental condition, we intercept all MotivGraph API calls and return complete texts or abstracts from the corpus of research papers used to build the knowledge graph. The researcher agent's access to knowledge is thus limited to this simulated interface, which provides document-level outputs rather than structured graph relationships.

3. \textbf{SCI-PIP Graph W/O Mentor}: This variant replaced our MotivGraph with the ``concept-paper'' graph constructed in SciPIP \cite{wang2024scipip}. SciPIP's built-in retriever was used to retrieve relevant entities from its graph structure, which were then used for knowledge augmentation. 

4. \textbf{W/O Graph + W/O Mentor}: In this variant, neither the MotivGraph nor the mentor agent's deliberation process was utilized. 

5. \textbf{W/O Semantic Scholar}: This variant retained the Semantic Scholar API for metadata retrieval but constrained its output to paper titles only, rather than complete metadata(including title, abstract, author, and publication year).

\subsection{Evaluation Setup}

Given the time-consuming and subjective nature of manual evaluation, and the documented efficacy of LLMs in judging text quality \cite{zheng2023judging, fu2023gptscore, liu2023g}, we adopted a model-based evaluation approach. This includes LLM direct evaluation and Swiss Tournament evaluation. For diversity assessment, we calculate diversity as \textbf{1-MeanSimilarity} among multiple ideas generated for the same topic \cite{si2024can}. See the Appendix \ref{Detailed Evaluation Methodology} for details.

\definecolor{lightblue}{rgb}{0.9, 0.95, 1} 
\vspace{-0.6em}
\begin{table}[htbp] 
    \centering 
    \small
    \renewcommand{\arraystretch}{1.3} 

    \begin{tabular}{|p{0.3cm}|l|c|c|c|>{\centering\arraybackslash}p{1.2cm}|} 
        \hline 
        \multicolumn{2}{|c|}{} & Nov. & Moti. & Exp. & Average \\ 
        \hline 

        \multirow{7}{*}{\raisebox{-0.5em}{\rotatebox[origin=c]{90}{Model Evaluation}}} 
        & \cellcolor{lightblue}Ours & \cellcolor{lightblue} \textbf{1072} & \cellcolor{lightblue} \textbf{1061} & \cellcolor{lightblue} \textbf{1061} & \cellcolor{lightblue} \textbf{1064} \\
        \cline{2-6} 

        & AI-Scientist-v2 &  1034 &  1016 &  1028 & 
        1026 \\ 
        \cline{2-6}

        & ResearchAgent & 1002 & 1011 & 1002 & 1005 \\
        \cline{2-6}

        & AI-Researcher & 1012 & 995 & 1001 & 1003 \\
        \cline{2-6}
        & RealPaper & 980 & 1020 & 1004 & 1001 \\
        \cline{2-6}
        
       & SciPIP & 1018 & 982 & 1002 & 1000 \\
        \cline{2-6}

        & CycleResearcher & 879 & 912 & 899 & 897 \\     
        \hline 

        \multirow{7}{*}{\raisebox{-0.5em}{\rotatebox[origin=c]{90}{Human Evaluation}}} 
        & \cellcolor{lightblue}Ours & \cellcolor{lightblue} 1038 & \cellcolor{lightblue} 1024 & \cellcolor{lightblue} 1026 & \cellcolor{lightblue} 1029 \\
        
        \cline{2-6}
        
        & RealPaper & \textbf{1071} & \textbf{1064} & \textbf{1063} & \textbf{1066} \\
        
        \cline{2-6}

        & AI-Researcher & 1013 & 1015 & 1020 & 1016 \\ 
        \cline{2-6}
        
        & AI-Scientist-v2 & 1010 & 1005 & 1013 & 1009 \\ 
        \cline{2-6}
      & ResearchAgent & 990 & 1003 & 1012 & 1002 \\
        \cline{2-6}
        & SciPIP & 1008 & 987 & 977 & 991 \\

        \cline{2-6}

        & CycleResearcher & 966 & 988 & 983 & 979 \\
        
        \hline 

    \end{tabular}
    \caption{Comparison of Ideation Methods}
    \label{tab:evaluation_comparison}
\end{table}
\vspace{-0.6em}

\subsection{Implement}

We selected three models—Qwen2.5-7B-Instruct \cite{qwen2025qwen25technicalreport}, DeepSeek-V3, and DeepSeek-R1 \cite{guo2025deepseek}—to investigate how models with different capabilities affect idea generation methods. Using a dataset of topics extracted from papers accepted at ICLR 2025, we generated at least three ideas per topic with each technique. Subsequently, we calculated the diversity of the generated ideas and employed Fast-Reviewer to quickly evaluate these ideas based on three dimensions: Novelty (Nov.), Experiment (Exp.), and Motivation (Moti.).

Additionally, we use DeepSeek-V3 to conduct a Swiss Tournament evaluation on the generated ideas across the Novelty, Motivation, and Experiment dimensions, computing ELO scores for each dimension and an overall average score. 

To further ensure the reliability of our evaluation, we replaced the automated Swiss Tournament assessment and LLM assessment with manual evaluations and reported corresponding ELO scores with direct scores. Since manual evaluation is time-consuming and labour-intensive, we only selected ideas generated by DeepSeek-V3, chosen topics, and selected one idea per topic for manual evaluation.

\section{Result and Analysis}
\subsection{Comparative Baselines}
Table \ref{main_table} presents the comparative results with the baselines. Experimental results show that our method has obvious advantages when DeepSeek-V3 generates ideas. Regarding diversity, Novelty, Experiment, and Motivation, our process is 0.03, 0.78, 0.25, and 0.49, higher than the second-best baseline regarding automatic evaluation. Manual evaluation results show that our method is 0.05, 0.05, and 0.25 higher than the second-best baseline regarding Novelty, Experiment, and Motivation. When the Qwen2.5-7B small parameter model is used, the model's ability to call APIs and integrate API return information is insufficient, and the number of API calls is abnormally high or low. At the same time, the context length that the small model can use is inadequate. In multiple rounds of modifications, part of the historical records often need to be discarded, which reduces the quality of idea generation to a certain extent. As for DeepSeek-R1, we can see that the Novelty and Motivation scores of the idea are still high due to the existence of the graph, but the scores of the three dimensions are lower than those of DeepSeek-V3. This is because the reasoning model requires long thinking, so the API call is planned before the API returns the result, which hinders the model from gradually exploring in depth.

Table \ref{tab:evaluation_comparison} compares the ELO score with the baseline. The results show that our method scores 28 points, 45 points, and 33 points higher than the second-best method (except Real Paper) in novelty, motivation, and experiments, respectively, and an average score of 38 points higher. The ELO scores of human evaluation are 25 points, 9 points, and 6 points higher in Novelty, Experiment, and Motivation, respectively.
\renewcommand{\arraystretch}{1.0} 
\setlength\extrarowheight{2pt}
\begin{table}[htbp]
 \centering
 \label{tab:ablation_study}
 \small 
\begin{tabular}{lccc}
 \toprule
 \textbf{Methods} & \textbf{Nov.} & \textbf{Exp.} & \textbf{Moti.} \\
 \midrule
 Ours & \textbf{8.39}/\textbf{6.45} & \textbf{8.64}/\textbf{6.7} & \textbf{8.70}/\textbf{6.7} \\
 \hdashline
 - w/o graph & 8.00/5.70 & 8.36/6.15 & 8.68/5.70 \\
 - w/ scipip-graph & 8.13/5.75 & 8.44/6.15 & 8.60/6.05 \\
 \hdashline
 - w/o mentor & 7.45/5.70 & 7.76/5.50 & 8.08/5.7 \\
 - w/o mentor \& graph & 7.71/5.65 & 8.19/5.70 & 8.47/5.90 \\
 \hdashline
 - w/o semantic scholar & 8.08/6.00 & 8.36/6.35 & 8.70/6.30 \\
 \bottomrule
    \end{tabular}
\caption{Results of ablation study on references and entities. The scores on the left of ``/'' are obtained using Fast-Reviewer evaluation, and those on the right are obtained by manual evaluation.}
\label{ablation}
\end{table}

\begin{figure}
    \centering
    \includegraphics[width=1.0\linewidth]{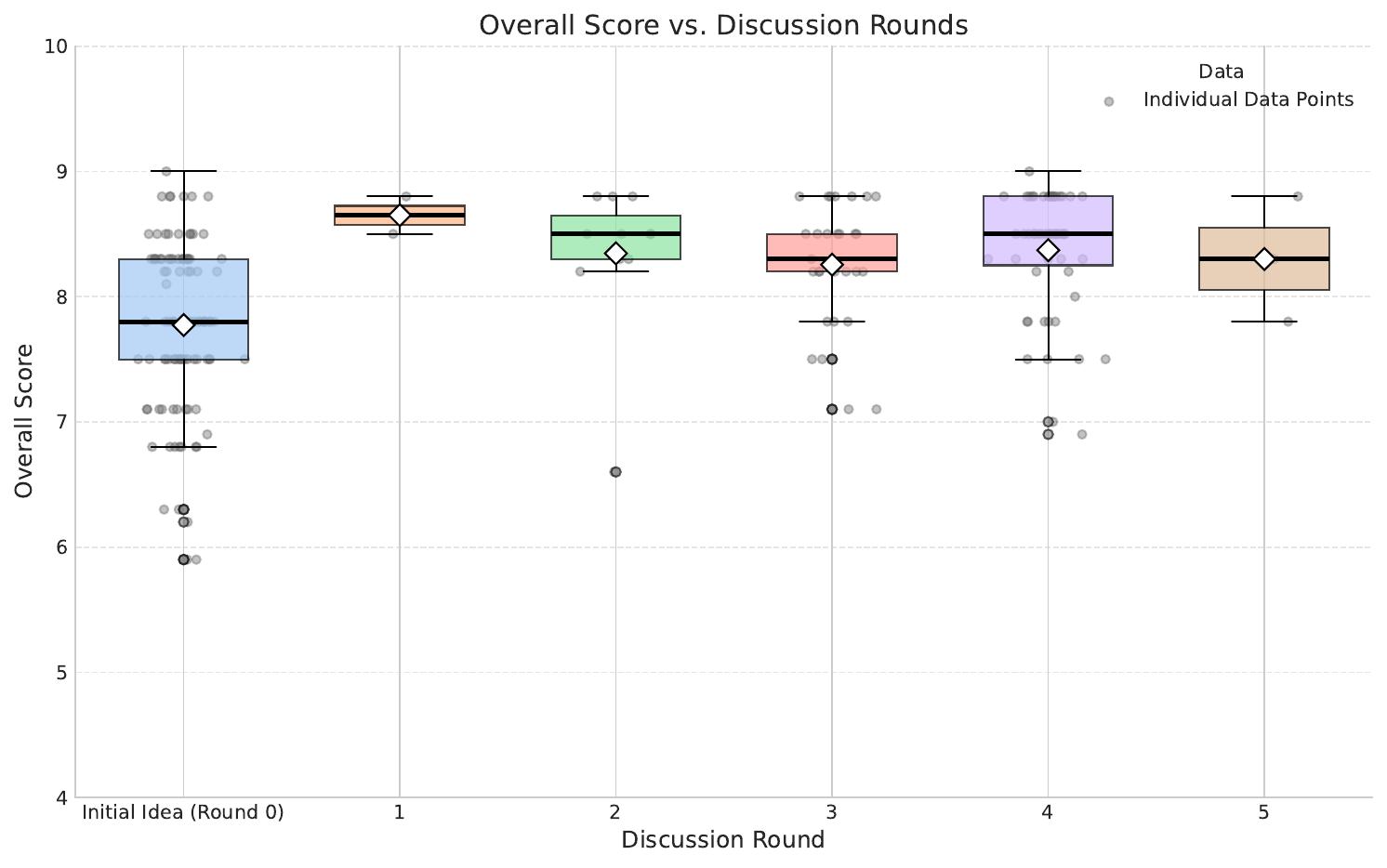}
    \caption{Final Score vs. Number of Discussion Rounds Plot}
    \label{fig:score_diff_vs_rounds}
\end{figure}

\subsection{Ablation Studies}
We performed a series of ablation studies to quantify the impact of key components within our proposed framework. Table \ref{ablation} shows the result.

First, we investigated the role of our designed knowledge graph. When we removed its hierarchical structure, relying solely on the raw source text, the Novelty, Experiment, and Motivation scores decreased by 0.39, 0.28, and 0.02 points, respectively. Replacing our graph with a generic 'scipip-graph' baseline also showed performance degradation, with scores dropping by 0.26 (Novelty), 0.20 (Experiment), and 0.10 (Motivation). These findings underscore the effectiveness of our specific knowledge graph design in boosting the innovation, feasibility, and underlying motivation of generated ideas.

Next, we examined the contribution of the mentor interaction phase. Ablating this step resulted in substantial decreases across all metrics: Novelty (-0.86), Experiment (-0.88), and Motivation (-0.62). This indicates that engaging in discussion and revision with a mentor improves the overall quality of generated ideas.

An interesting observation was made when the knowledge graph was turned off in addition to the mentor interaction (w/o mentor + w/o graph condition). In this setup, scores were higher than in the w/o mentor condition alone. We hypothesise that this phenomenon occurs because the model generates more conventional ideas without the graph introducing potentially divergent nodes that might achieve a higher initial score without expert guidance. Figure \ref{fig:score_diff_vs_rounds} shows the final score versus the number of discussion rounds.

Finally, we assessed the contribution of the detailed paper information from Semantic Scholar. Removing all semantic content except for the title led to decreases in both Novelty and Experiment scores. This suggests that the comprehensive background knowledge from Semantic Scholar is beneficial for generating innovative and experimentally grounded ideas.

\begin{figure}
    \centering
    \includegraphics[width=1.0\linewidth]{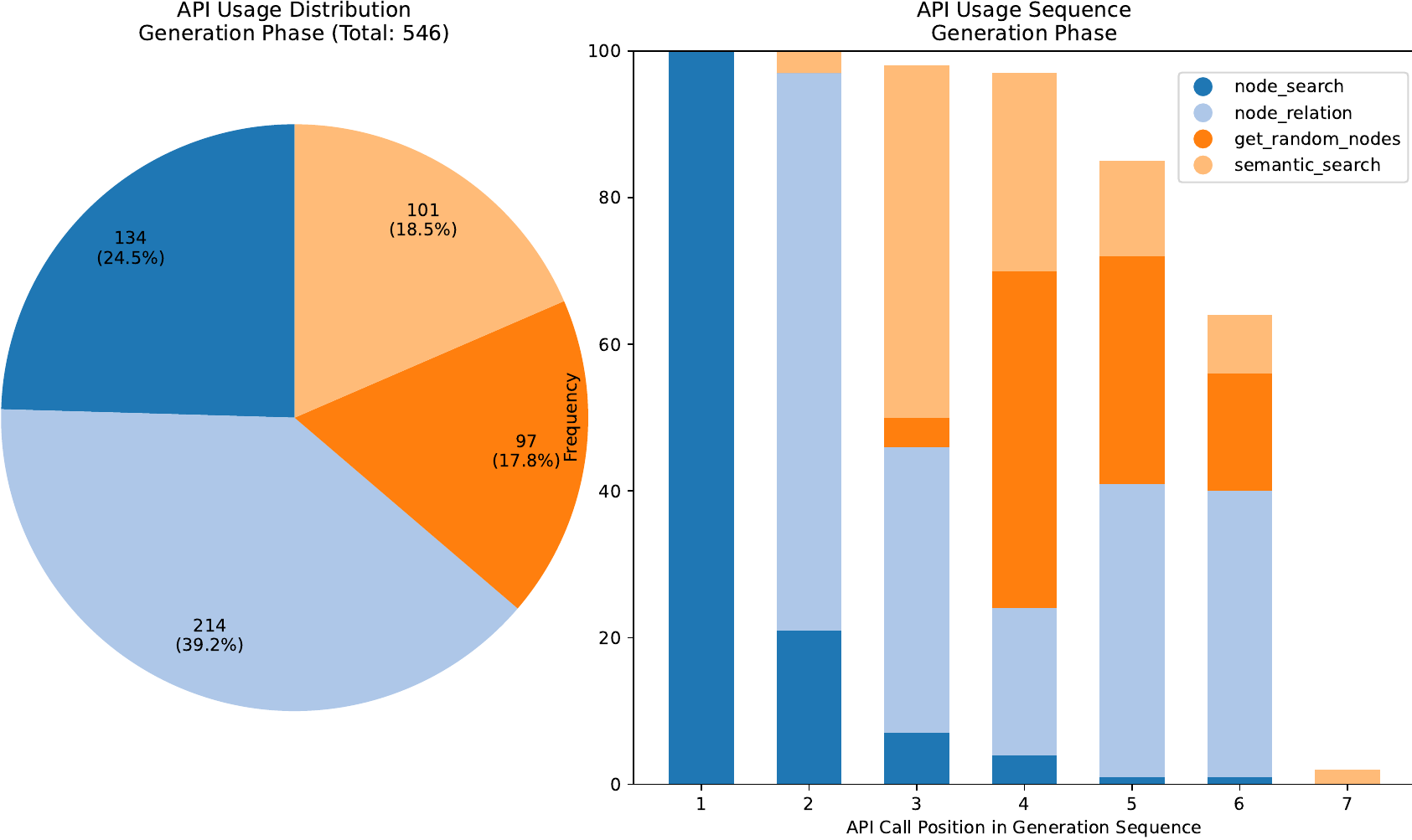}
    \caption{researcher agent API usage and sequence in ideation}
    \label{fig:generate_api_usage}
\end{figure}

\subsection{Further Analysis}
This subsection discusses the intermediate results produced during the idea generation process.
\paragraph{Idea score vs. the number of rounds.}
Figure \ref{fig:score_diff_vs_rounds} illustrates the relationship between the final idea score and the number of discussion rounds. From this figure, it can be observed that discussion contributes to an improvement in overall quality. Furthermore, a higher initial quality often correlates with fewer discussion rounds, and scores are notably higher when the mentor raises fewer questions. Nevertheless, engaging in more discussion rounds can also enhance the overall quality of the ideas.

\paragraph{API Usage.}
Figures \ref{fig:generate_api_usage} present the frequency and distribution of API calls made by the researcher agent across different rounds during the idea generation process, respectively. These figures demonstrate that our constructed researcher agent can autonomously invoke tools and independently determine tool usage based on the specific problem context.

\paragraph{Differences between backbone models.}
As shown in the table \ref{tab:llm-evaluator-qwen}, for the Qwen model, increasing the parameter count from 7 B to 32 B yields a marked improvement in idea quality. Overall, our method's performance can benefit from stronger model capabilities; however, when the model size reaches 72 B, quality actually declines. Our observations reveal that Qwen-2.5-72 B begins to produce garbled output under long-context conditions, which we believe indicates a sharp drop in its comprehension and reasoning capabilities once the input exceeds a certain length. We observed the same behavior with Qwen3. Indeed, Qwen3 generated extensive mixed-language garble that prevented the pipeline from functioning correctly. Consequently, we conclude that Qwen models show a substantial performance gap compared to DeepSeek when tasked with understanding and analyzing large volumes of text.

For the analysis experiment on 'Generalizability to Other Scientific Domains,' please see the appendix \ref{Generalizability to Other Scientific Domains.}.

\renewcommand{\arraystretch}{1.0} 
\setlength\extrarowheight{2pt}
\begin{table}[htbp]
 \centering
 \label{tab:ablation_study}
 \small 
    \begin{tabular}{lccc}
      \toprule
      \textbf{Model}       & \textbf{Novelty} & \textbf{Motivation} & \textbf{Experiment} \\
      \midrule
      DeepSeek-V3          & 8.39             & 8.70                & 8.64               \\
      Qwen-7B        & 6.46             & 6.52                & 6.44               \\
      Qwen-14B       & 6.55             & 6.95                & 7.33               \\
      Qwen-32B       & 6.85             & 7.95                & 7.70               \\
      Qwen-72B       & 6.90             & 7.05                & 6.55               \\
      \bottomrule
    \end{tabular}%
  \caption{LLM‐evaluator scores for different Qwen model sizes and DeepSeek‐V3.}
  \label{tab:llm-evaluator-qwen}
\end{table}

\section{Conclusion}
LLMs offer great promise for academic ideation but face challenges with idea grounding and confirmation bias. We introduce MotivGraph-SoIQ, a novel framework that enhances LLM ideation by integrating a Motivational Knowledge Graph for grounding from literature and a Q-Driven Socratic Ideator. This dual-agent system uses Socratic questioning to refine ideas, mitigating confirmation bias and improving novelty, experimental feasibility, and motivation. Our results demonstrate MotivGraph-SoIQ's effectiveness and superior performance across LLM-based scoring, ELO ranking, and human evaluation. Ablation studies confirm the crucial contributions of both MotivGraph and the Socratic dialogue. This work highlights the power of combining structured knowledge with interactive, critique-based refinement for robust LLM ideation.

\section{Limitations}
While our findings are promising, we acknowledge several limitations in the current work. The scope of our constructed MotivGraph is presently limited, primarily encompassing knowledge within the AI domain and lacking comprehensive coverage of other scientific disciplines. Expanding its domain coverage is essential for realising the full potential of cross-disciplinary idea generation. However, our constructed MotivGraph holds considerable potential for uncovering connections across diverse scientific disciplines and presenting these associations to large language models for their utilisation. Furthermore, due to constraints on available resources and time, our experimental validation was conducted on a specific dataset size, and we evaluated the framework using a limited variety of LLM models. Future work should focus on scaling up the experimental evaluation to a larger dataset and testing a more diverse range of underlying LLMs to confirm the generalizability of our findings.

For future research, we also plan to explore extending the MotivGraph to incorporate other academic knowledge and relationships. Further investigation into alternative dialogue strategies within the Socratic framework could yield additional insights.

\section{Ethics Statement}

Our system is developed with the explicit and sole purpose of serving as an assistive tool to augment human creativity and facilitate the discovery of novel research ideas within the academic domain. Our goal is to empower researchers by providing inspiration, helping to overcome ideation blocks, and suggesting potentially fruitful avenues for investigation grounded in existing knowledge.

We unequivocally condemn and strongly disavow any potential misuse of this system. This includes, but is not limited to, using the system to generate ideas or methods for illegal activities, unethical research practices, harmful technologies, malicious applications, or any purpose that could cause societal harm, violate privacy, or infringe upon human rights. Users are solely responsible for the evaluation, validation, and ethical implications of any system-generated idea and its subsequent application. The system is designed to be a creative aid, not an autonomous decision-maker or a substitute for human ethical reasoning and responsibility.

\section{Acknowledgement}

This work is supported by the National Natural Science Foundation of China (No.U24A20335, No. 62176257, No.62576340). This work is sponsored by Beijing Nova Program (No.20250484750) and supported by Beijing Natural Science Foundation (L243006). This work is also supported by the Youth Innovation Promotion Association CAS.

\bibliography{custom}
\bibliographystyle{acl_natbib}

\appendix
\section{Generalizability to Other Scientific Domains.}
\label{Generalizability to Other Scientific Domains.}
Theoretically, our method, MotivGraph-SoIQ, offers strong generalizability across disciplines. Its MotivGraph component supplies the large model with a motivational foundation for idea conception in <Problem, Challenge, Solution>, reflecting a basic scientific‐research paradigm in many fields. Moreover, using Socratic dialogue to refine ideas iteratively is likewise a common research practice.

We collected 185 recent papers from high-quality medical journals (Nature Medicine, Nature Biomedical Engineering, and IEEE Transactions on Medical Imaging) to validate our approach in another domain empirically. We clustered these into 30 topics for idea generation and used the remaining 155 papers to construct a small‐scale knowledge graph. We then compared our method against two strong baselines: ResearchAgent and CycleResearcher (a domain-knowledge fine-tuned model). Because our original evaluator was trained only on AI‐domain papers, we replaced it here with DeepSeek-r1, which offers comparable performance. Table \ref{tab:llm-evaluator} shows that MotivGraph-SoIQ continues to perform effectively in the medical domain.
\renewcommand{\arraystretch}{1.0} 
\setlength\extrarowheight{2pt}
\begin{table}[htbp]
 \centering
 \label{tab:ablation_study}
 \small 
    \begin{tabular}{lccc}
      \toprule
      \textbf{Model}       & \textbf{Novelty} & \textbf{Motivation} & \textbf{Experiment} \\
      \midrule
      Ours                  & 8.04            & 8.72               & 8.02               \\
      ResearchAgent         & 7.55            & 8.27               & 8.01               \\
      CycleResearcher       & 6.85            & 7.87               & 6.89               \\
      \bottomrule
    \end{tabular}%
  \caption{LLM‐evaluator scores for different methods.}
  \label{tab:llm-evaluator}
\end{table}
\section{Details}

\subsection{Baseline implement}
\label{baseline}
\paragraph{AI-Researcher:}
This method represents a simple yet effective LLM ideation approach that integrates Retrieval-Augmented Generation (RAG), filters duplicate ideas using vector similarity, and employs an LLM-based automatic ranker inspired by a Swiss-tournament design. It is a typical example of using a single LLM agent for idea generation without explicitly constructing a knowledge graph. Comparing with AI-Researcher helps us understand the performance level of a general or relatively straightforward LLM generation agent in the context of ideation. The ideas it generates primarily include ``Motivation'', ``Proposed Method'', ``Step-by-Step Experiment Plan,'' and ``Test Case Examples,'' which share similarities in structure with the ideas produced by our method.

\paragraph{Cycle Researcher:}
This method introduces Iterative Preference Training, leveraging extensive prior literature and review feedback to train the Cycle Researcher model. It can generate paper proposals covering motivation, idea (method), and experimental setup, a structure akin to our generated ideas. We use Cycle Researcher as a baseline to assess the ideation capability of LLMs trained through reinforcement learning. In our experiments, we opted for the 12B model for comparison primarily to conserve idea generation time and resources. As indicated in the original Cycle Researcher paper, the 12B model exhibited performance comparable to, and in many metrics even superior to, their 123B model for this task. Additionally, we replaced their original Bib literature database with the Semantic Scholar API for the RAG component.

This is a section in the appendix.

\paragraph{AI-Scientist-v2:}
This is an improved version of AI-Scientist \cite{yamada2025ai}, which enriches the content of generated ideas and integrates Semantic Scholar as a function call. Similar to our method's approach to external information retrieval, AI-Scientist-v2 utilizes external knowledge via API calls. In generating ideas for comparison using AI-Scientist-v2, we set the number of reflection steps to 5 and generated five ideas per topic.

\paragraph{SciPIP:}
The core methodology of SciPIP \cite{wang2024scipip} lies in combining multi-angle literature retrieval and a dual-path idea generation strategy. It first retrieves literature content and related entities based on the provided topic and then generates ideas through brainstorming and RAG. Similar to our method, SciPIP constructs a knowledge graph, specifically a ``concept-paper'' graph where concepts are extracted from papers by a large model. However, it does not explicitly structure knowledge around challenges and solutions. This method serves as an excellent comparison point to demonstrate the effectiveness of our Challenge-Solution Knowledge Graph. We used SciPIP for standalone idea generation and integrated its graph entity retrieval module to replace our Challenge-Solution Knowledge Graph during the idea generation process to compare the graph structures directly.
We use the dual-path approach mentioned in the paper for idea generation.

\paragraph{ResearchAgent:}
ResearchAgent \cite{baek2024researchagent} is a system designed to assist researchers in iterative research idea generation using Large Language Models (LLMs). It aims to produce novel and impactful research ideas by augmenting information from scientific literature and employing collaborative LLM-driven review agents for iterative optimization. Its strategy of information enhancement via an ``academic graph + entity knowledge storage'' and iterative optimization through a multi-agent collaborative review loop shares similarities with our method but presents distinct differences, making it a valuable subject for comparative experiments.

\subsection{Fast-Reviewer Test:}
We tested Fast-Reviewer Test on our constructed dataset, measuring AUC scores for positive-negative discrimination across Novelty, Experiment, and Motivation categories.
Table \ref{auc_evaluation} shows the AUC scores.

\begin{table}[htbp] 
    \centering 

    \begin{tabular}{l *{3}{S[table-format=0.2]}}
        \toprule 
        \multirow{2}{*}{Model}
        & \multicolumn{3}{c}{AUC} \\
        \cmidrule(lr){2-4} 
        & {Novelty} & {Motivation} & {Experiment} \\
        \midrule 

        Fast-Reviewer & \textbf{0.76} & 0.56 & 0.66 \\
        DeepSeek-V3   & 0.68 & 0.57 & 0.53 \\
        deepseek-R1   & 0.75 & 0.58 & \textbf{0.70} \\
        \bottomrule 
    \end{tabular}

    \caption{AUC Score} 
    \label{auc_evaluation} 

\end{table}

\subsection{SciMotivMiner rules:}
\label{SciMotivMiner rules}
\paragraph*{Problem Entity ($P_i$):}
The name of the \textbf{Problem} entity ($P_i$) represents the overall research task, domain, or high-level objective that the paper addresses. The naming follows the structure `[General Task/Field of Study]` and aims for \textbf{3-7 words}. These names must be generalized and \textbf{strictly avoid} authors' specific, non-generalized names or abbreviations. Problem entity names are derived solely from the context in the paper's Introduction section.

The corresponding problem description provides a brief (ideally \textbf{1-2 sentences}), neutral, high-level definition of the overall research task, field of study, or objective. This description focuses solely on the task or field or its general purpose/goals, presented as a standalone concept. Crucially, this description \textbf{must not} include any mention of challenges, limitations, difficulties, or specific areas of focus motivated by these challenges. It is formulated as a universal definition, informed by the Introduction section.

\paragraph*{Challenge Entity ($C_i$):}
The name of the \textbf{Challenge} entity ($C_i$) captures a specific, atomic difficulty, limitation, gap, or existing shortcoming \textit{within the identified Problem} that the paper aims to address. Its naming strictly adheres to the structure `[Specific Difficulty/Limitation] in [Aspect of Problem/Domain Context]` to clearly state the precise difficulty and its context within the problem or domain. These names aim for \textbf{5-8 words}, prioritizing the required structure and specificity. As with problem entities, authors' specific, non-generalized names or abbreviations for challenges are \textbf{strictly avoided}, and names are derived from the Introduction section.

The challenge description, summarized from the Introduction section (ideally \textbf{2-3 sentences}), explains this specific difficulty, limitation, or gap and details how it relates to the broader Problem.

\paragraph*{Solution Entity ($S_i$):}
For the \textbf{Solution} entity ($S_i$), the name captures the essential technical approach, category, or fundamental principle employed to address the Challenge. A crucial constraint is that the authors' specific name, acronym, or code name for their proposed solution (or any non-generalized term they introduce) is \textbf{strictly not used} in the entity name, drawing instead on general technical terms or descriptions of the solution's core components or principles. Solution names aim for \textbf{7-10 words}. For solutions described with a citation in the Introduction, their established general name or common abbreviation (if widely recognized and within the word count aim) is used, based on the Introduction description. For novel solutions (typically described without a citation in the Introduction), the solution section is consulted to understand the core technical approach and fundamental principles, and the name is generated using general technical terms or essential component descriptions based on this technical understanding from both sections.

The solution description provides a brief (ideally \textbf{2-3 sentences}) explanation of the solution's core technical aspects, focusing on \textit{how} it works technically. If the Introduction's description is high-level, results-focused, or lacks sufficient technical detail, the solution section is consulted to incorporate key technical aspects explaining the approach.

\subsection{Detailed Hierarchical Parent Node Addition}
\label{Hierarchical Parent Node Addition}
Following the extraction process from the papers, we obtain a set of $n$ distinct knowledge triplets, ${(P_i, C_i, S_i)\}_{i=1}^n}$. While initially extracted as independent triples, the problems and challenges described within them exhibit inherent relationships. Academic problems inherently possess a hierarchical structure, and different papers address problems at varying granularities. For example, one paper might focus on the broad area of 'Machine Translation', while another delves into 'Low-Resource Machine Translation for Indigenous Languages'. To capture these relationships and further associate the knowledge, we construct a hierarchical structure for the graph by introducing parent nodes for both Problem ($P$) and Challenge ($C$) entities. This hierarchical organisation is crucial to prevent the knowledge base from becoming overly fragmented or unstructured, making it challenging to comprehend and navigate. Without this hierarchy, the graph would fail to fully leverage its advantages for thoroughly organizing complex information, hindering compelling exploration during subsequent ideation processes.

To acquire these parent nodes and establish hierarchical relationships within the sets of Problem ($P$) and Challenge ($C$) nodes, we propose the \textbf{Parent Node Addition Algorithm}. This process is applied separately to the Problem ($P$) and Challenge ($C$) node collections.

All original Problem and Challenge nodes are initially embedded into a vector space to enable subsequent similarity search based on their semantic representations. This vector space representation is fundamental for quantifying the semantic relationships between nodes.

The algorithm operates on an initial set $S$, which at the start of the process, contains all nodes from either the Problem or Challenge set being processed. The core mechanism involves iteratively processing nodes within this set $S$ until it becomes empty.

The algorithm maintains $S$ as a dynamic working set. It repeatedly selects a node $N$ from the current set $S$. For this focal node $N$, the algorithm identifies its $k$ most similar neighbours based on the pre-calculated vector embeddings. A critical filtering step is then applied: only those similar neighbors that \textit{also} remain present in the current working set $S$ are retained as potential candidates for grouping with $N$. Let this filtered set of eligible similar nodes be $V_{filtered}$.

A Large Language Model (LLM) is crucial at this stage. It evaluates the semantic coherence and potential for forming a higher-level concept when considering the focal node $N$ and the nodes in $V_{filtered}$. Based on this evaluation, the LLM decides whether a merge operation should occur.

A new parent node is created if the LLM determines that a merge is appropriate and the set $V_{filtered}$ is not empty. This new node represents a more general theme or domain that encapsulates the concepts expressed by $N$ and the nodes in $V_{filtered}$. Directed edges, labelled \textit{parent-of}, are added from this new parent node to $N$ and to every node $v \in V_{filtered}$, establishing their hierarchical link.

Following the decision and potential merge, the current node $N$ is removed from the set $S$, as it has been processed in this iteration. Furthermore, if a merge occurred, all the nodes in $V_{filtered}$ that became children of the new parent node are also removed from the set $S$. This dynamic update ensures that each node is considered for forming a parent at most once in this pass and that nodes already integrated into a higher level via merging are no longer candidates within the same pass.

The iterative selection and processing of nodes from the set $S$ continues until $S$ becomes empty. At this point, all nodes from the initial set have been either processed as a focal node or removed because they were merged as children. The parent nodes created during this process represent a higher level of abstraction for the grouped concepts within the original set $S$.

\subsection{Dataset Construction and Evaluation Details}
\label{Dataset Construction and Evaluation Details}
MotivGraph Dataset
We constructed the MotivGraph from 8625 accepted papers from ICLR 2024, ICML 2024, and NeurIPS 2024, collected from OpenReview and other sources. Using the SciMotivMiner method (detailed in Section \ref{MotivGraph construction}) with DeepSeek-V3 as the extractor on the full text of these papers, we obtained 25515 solution nodes, 31158 challenge nodes, and 12137 problem nodes. Node descriptions were vectorized using all-MiniLM-L6-v2 \cite{reimers2019sentence}. Subsequently, the Hierarchical Parent Node Addition method (detailed in Section \ref{Hierarchical Parent Node Addition}) established 37367 PARENT\_OF relationships and added 7089 parent nodes.

Evaluation Dataset
We clustered the titles of all accepted ICLR 2025 papers for the evaluation dataset using all-MiniLM-L6-v2. From these clusters, we selected 100 papers(excluding papers used for Fast-Reviewer training) representing diverse topics. DeepSeek-V3 \cite{liu2024deepseek} extracted each selected paper's core idea and topic. The extracted core ideas served as ground truth, matching our method's output format for subsequent comparisons, while the extracted topics served as input for the idea generation process.

\subsection{Detailed Researcher Agent's Toolset}
\label{Detailed Researcher Agent's Toolset}
The researcher agent within our Q-Driven Socratic ideator has four specialized tools to facilitate comprehensive knowledge exploration and foster innovative ideation.

\subsection*{Graph Node Fuzzy Search}
The researcher agent provides a search query. This API returns the names and types of the top K similar nodes based on the semantic similarity between the search query and the descriptions of nodes in the Motivational Knowledge Graph (MotivGraph). This tool enables the researcher Agent to gain an overarching understanding of the target domain or task by identifying related \textit{problem}, \textit{challenge}, and \textit{solution} entities within that domain.

\subsection*{Graph Node Relation Retrieval}
Given the name of an interesting node, this API returns the node's description, names, and types of its neighboring nodes. The researcher agent can retrieve hierarchical relationships between nodes and the (problem, challenge, solution) triplets representing critical motivational information through this tool. This contextual information deepens the researcher agent's understanding of the target domain/task, facilitates more effective subsequent retrieval, and establishes a robust foundation for the ideation phase.

\subsection*{Semantic Scholar Literature Search}
The researcher agent provides a search query to this API, which returns relevant academic literature. In contrast to the structured knowledge supplied by the MotivGraph, Semantic Scholar offers more specific and granular information, allowing the ideator agent to understand particular challenges or technologies comprehensively.

\subsection*{Get Random Nodes to Enhance Novelty}
The researcher agent autonomously enters the ideation phase after sufficient knowledge exploration and comprehensively understands the target domain or task. During this phase, the random\_nodes API obtains disparate, randomly selected nodes. The researcher agent's primary objective is to leverage its domain understanding and attempt to apply these obtained random nodes (which can include \textit{problem}, \textit{challenge}, and \textit{solution} entities) to the target domain or task. This involves seeking potential connections, adaptations, or insightful modifications.

This process directly supports the "creativity-relevant processes" component of Amabile's Componential Theory of Creativity \cite{amabile1996creativity}, which posits that motivation constitutes one of the three essential components of innovation (alongside domain-relevant skills and creativity-relevant processes), with intrinsic motivation being particularly critical for breakthrough ideation. This mechanism is vital for ensuring the novelty of the generated ideas. Simultaneously, the researcher agent, equipped with sufficient knowledge ("domain-relevant skills"), particularly after internalizing the motivation encoded in the (problem, challenge, solution) triples, benefits from the inherent logical progression within this motivational information (i.e., research domain/task → specific challenges → solutions addressing challenges). This inherent logic can foster "intrinsic motivation" within the ideator agent. Driven by this intrinsic motivation, the researcher agent attempts to adapt the external (random) nodes to the target domain, aiming to identify potentially new challenges within the target domain based on external challenges, or to discover novel ways to solve a target challenge by adapting external solution concepts.
\subsection{Detailed Evaluation Methodology}
\label{Detailed Evaluation Methodology}
We employed a multifaceted model-based evaluation strategy to assess the quality of generated research ideas. This approach can evaluate the quality of ideas holistically without using time-consuming and labor-intensive manual evaluation, leveraging recent advancements in LLM judgment capabilities\cite{zheng2023judging}.

LLM Direct Evaluation (Fast-Reviewer)
We fine-tuned Fast-Reviewer, an LLM specifically for direct idea quality assessment. This model was trained on a dataset derived from ICLR 2025 OpenReview comments. We utilised Qwen2.5-7 B-Instruct to extract positive and negative labels for novelty, experimental soundness, and motivation from 1200 training papers and 287 test papers. Additionally, DeepSeek-V3 was used to extract core ideas from these papers. Finally, Qwen2.5-7 B-Instruct was fine-tuned to this dataset to create a Fast-Reviewer. As shown in Table \ref{auc_evaluation}, Fast-Reviewer achieves evaluation capabilities similar to deepseek-r1 but with lower cost and faster inference.

Swiss Tournament Evaluation
Following established pairwise comparison methodologies like the Swiss tournament \cite{si2024can} and Idea Arena \cite{li2024chain}, we implemented a Swiss Tournament Evaluation. Different idea generation methods competed in a series of rounds for each topic. An LLM performed pairwise judgments on the quality of ideas, and these outcomes updated the ELO scores for each method. The final ELO scores provided an unbiased estimate of their relative performance. This method addresses concerns regarding LLMs' insufficient diversity in idea generation \cite{si2024can}.

\section{Case Study:}
To illustrate the gap between the ideas generated by our proposed method and high-quality ideas from authentic papers, we present the following case study in Figure \ref{Case}:

This is the method description for the idea our approach generated on the topic ``LLM‐based agent security: Benchmarking attacks and defenses in LLM‐based agents.'' The proposed idea introduces representation trajectory analysis from dynamical systems theory, tracking the model's hidden‐layer activations to detect whether it remains in a ``normal'' state, and quantifies security‐critical failures (e.g., task hijacking, privilege escalation, or data leakage) by measuring the Minimum Variation Distance (MVD): the smallest prompt perturbation strength needed to induce such failures. Finally, it defines an Agent Vulnerability Index (AVI), which systematically dissects the agent's architecture (including model and component code) through controlled component removal or modification, revealing each component's impact on overall security performance.

On the surface, this appears promising, by altering inputs, one can observe when the agent drifts toward unsafe outputs. However, the proposed prompt‐perturbation scheme lacks a principled design: realistically, breaching a large model or its composed agent system typically requires carefully engineered attacks (e.g., inserting invisible or non‐standard characters), not mere lexical substitutions. Moreover, the representation‐trajectory approach is hard to apply in practice. Given the opaque internal mechanics of large models, it is difficult to infer an ongoing attack or security breach solely from hidden‐state trajectories, thus determining the model's safety status. The AVI metric likewise proves challenging to compute: agent components are often tightly coupled, so removing one component may render the system inoperative, preventing meaningful measurement.

In summary, this case study shows that while our method can pinpoint innovative angles relevant to the topic and generate coherent ideas, it lacks additional domain expertise and research experience in designing core attack and defense techniques, leading to feasibility gaps. Future work should enhance the agent's domain knowledge and research experience. Nonetheless, although our generated ideas still fall short of the immediately actionable, high‐quality proposals extracted from authentic ICLR papers, they exhibit strong logical creativity. They can serve as valuable inspiration for human researchers.

\begin{figure}[htbp]
\centering
\begin{tcolorbox}[
    title=PSBench Methodology,
    fonttitle=\bfseries\small,
    colback=white,
    colframe=green!50!black,
    colbacktitle=green!10!white,
    coltitle=green!40!black,
    boxrule=1pt,
    arc=3mm,
    width=0.5\textwidth,
    bottom=5mm  
]
\doublespacing  

\textbf{\large 1. Variation Generation:}
\begin{itemize}
    \item[] Create 1000+ prompt variants per input using:
    \begin{itemize}
        \item[a)] Lexical transformations (synonyms, typos)
        \item[b)] Semantic paraphrasing (LLM-generated)
        \item[c)] Structural changes (instruction reordering)
    \end{itemize}
\end{itemize}

\medskip
\textbf{\large 2. Trajectory Instrumentation:}
\begin{itemize}
    \item[] Track internal states using:
    \begin{itemize}
        \item[a)] Hidden state snapshots every 3 layers
        \item[b)] Attention pattern logging
        \item[c)] Gradient flow analysis
    \end{itemize}
\end{itemize}

\medskip
\textbf{\large 3. Metric Computation:}
\begin{itemize}
    \item[] 
    \begin{itemize}
        \item[a)] MVD: Optimal transport distance to failure boundary
        \item[b)] TDS: Curvature analysis of state trajectories
        \item[c)] AVI: Architecture component ablation testing
    \end{itemize}
\end{itemize}

\medskip
\textbf{\large 4. Benchmark Suite:}
\begin{itemize}
    \item[] 
    \begin{itemize}
        \item[a)] Security scenario test cases
        \item[b)] Reference agent implementations
        \item[c)] Baseline comparison protocol
    \end{itemize}
\end{itemize}

\end{tcolorbox}
\caption{Idea generated by our method for the topic of ``LLM-based agent security: Benchmarking attacks and defenses in LLM-based agents''}
\label{Case}
\end{figure}

\section{Prompt:}

\subsection{API SELECT TEMPLATE}
\begin{lstlisting}[frame=lines]
# Tool Introduction: The following tools can help you complete your task.
1. Knowledge Graph: This graph consists of (Problem, Challenge, Method) triplets and parent problem and challenge nodes. Triplet pairs belonging to the same problem or challenge type are connected through the parent problem or challenge node.
Using this graph for ideation typically requires multiple API calls:
Three API tools help you work with the graph: node_search(), node_relation(), and get_random_node(). Below is a detailed introduction to these three APIs:

# API Tool Call Format: Output the following format. Importantly, be sure to output the special token: <CALL> at the end.
```function call
conducting function_name(parameter_name=
parameter_value)
special token: <CALL>
```

## node_search(search_query="<your content of interest>"):
- Function: node_search(search_query="<your content of interest>")
- Description: This API allows you to perform a fuzzy search for your content of interest. You will receive the names of nodes in the graph related to your search, including problems, challenges, and methods.
- Usage: By providing a search term (e.g., "LLM Compression"), you can retrieve the names of nodes related to that query.
- Use Example:
```function call
conducting node_search(entity_name_list="LLM Compression")
Special token: <CALL>
```

## node_relation(entity_name_list=
["<node name you're interested in>",...])
- Function: node_relation(entity_name_list)
- Description: This API allows you to retrieve detailed information about the nodes in the input list, including the nodes connected to it and the relationships between them.
- Usage: You can retrieve the node name using node_search(), then select the node of interest to explore using this API. You can continue exploring along a specific path.
- Example:
```function call
conducting node_relation(entity_name_list=["LLM Compression","DistilledLM"])
Special token: <CALL>
```

## get_random_nodes(number=10):
- Function: get_random_nodes(number=10)
- Description: This API allows you to retrieve 10 random nodes, including problem, challenge, and method. These nodes are the source of your innovation. You need to research and think about how to use these nodes for ideation. - Usage: get_random_nodes(number=10)
- Example:
```function call
conducting get_random_nodes(number=10)
Special token: <CALL>
```

2. Semantic Scholar: You can use this API to retrieve literature and deepen your understanding of a research topic.
semantic_search(search_query="<your interest>")
- Function: semantic_search(search_query="<your interest>")
- Description: You can use this API to query literature and find papers related to your search query, which can help you understand a field.
- Usage: Provide a search_query (e.g., "LLM Compression"). The API will return the titles and abstracts of the top 20 papers related to that query. The search_query must be in English. If the result is empty, please adjust your search_query or retry.
-Example:
```function call
conducting semantic_search(search_query="LLM Compression")
Special token:<CALL>
```

Note:<CALL> is a marker for calling functions. If this marker is not present, the function will not be called. Please ensure the special token is output correctly.
\end{lstlisting}

\subsection{IDEA GENERATION TEMPLATE}
\begin{lstlisting}[frame=lines]
You are an experienced AI researcher who aims to propose high-impact research ideas resembling exciting grant proposals. Feel free to suggest any novel ideas or experiments; make sure they are novel. Be very creative and think out of the box. Each proposal should stem from a simple and elegant question, observation, or hypothesis about the topic.
The IDEA JSON should include the following fields:
- "Name": A short descriptor of the idea. Lowercase, no spaces, underscores allowed.
- "Title": A catchy and informative title for the proposal.
- "Motivation": A single string describing the thought process that led to the conception of this idea. Articulate the rationale and context using fluent, academic language.(approximately 250 words).
- "Related Work": A section that introduces foundational work related to each core component of your idea, especially content related to new concepts you introduce. It should demonstrate the strengths and weaknesses of existing research related to your topic and highlight the innovation of your own research. Represent the paper from semantic_search() with a citation in the format of '(<author name here> et al., <year here>)'.
- "Abstract": An abstract that summarizes the proposal in conference format (approximately 250 words).
- "Method": A single string containing a detailed description of the entire method. This string should outline your method step-by-step, explaining the key procedures involved. Focus on providing a clear, comprehensive explanation of how your method works from beginning to end. Discuss why these steps are important and how they directly contribute to solving the problem addressed in the idea.
- "Experiments plan": A single string containing a detailed plan for experiments to validate the proposal. The description should outline the experiments to be conducted, ensuring they are simple and feasible. Be specific about how the hypothesis would be tested, detail any precise algorithmic changes, and include the evaluation metrics to be used. Explain the rationale behind conducting these experiments and how they would prove the effectiveness of each component of the proposed method.
- "Risk Factors and Limitations": A single string containing a description of the potential risks and limitations of the proposal. This string should discuss various potential risks that might hinder the successful implementation or outcome of the proposed idea, as well as inherent limitations of the approach.
For any of the above fields:
If you are inspired by entities from the Knowledge Graph, you should reference them using the <a href="...">...</a> hyperlink format. When using this method, indicate the entity name and entity type, as this approach helps to improve language fluency.
For example: "Despite Large Language Models demonstrate strong capabilities in automating text generation, they still face some inherent challenges when applied to tasks requiring creativity, such as research idea generation. A significant issue is that &lt;a href="kg:challenge:Suboptimal initial output generation in language models">the initial ideas generated by models are often repetitive and suboptimal&lt;/a>. This makes subsequent idea development and filtering more time-consuming."
Ensure the JSON is properly formatted for Automatic parsing. Please ensure the output strictly adheres to JSON format specifications: use double quotes for keys and string values, escape internal quotes with \", avoid trailing commas, and exclude non-JSON elements like comments or unquoted keys.

Output Format for the Idea:
IDEA JSON:
```json
{
"Name": "...",
"Title": "...",
"Motivation": "...",
"Related Work": "...",
"Abstract": "...",
"Method": "...",
"Experiments plan": "...",
"Risk Factors and Limitations": "..."
}
```
Here are some tools for you to use:
[TOOLS]

# Task: Complete the following three tasks in order, using only the ideas in the graph. Invoke the tools multiple times to output the final idea. Your research topic is: [TOPIC]
## Task 1: Understanding Your Research Task/Topic: Task Objective: Fully understand the problems, challenges, methods, and related literature related to your topic to lay a solid foundation for further exploration.
Output your Task 1 exploration results:

Task Thinking Guide: First, you need to use node_search() several times to identify problem, challenge, and method nodes in the knowledge graph that are relevant to your research. For the returned results, you can also use node_relation() several times to obtain detailed information about the nodes, including descriptions, relationships, and so on. You can also use semantic_search() to explore related literature to further strengthen your understanding of your research field.
## Task 2: Creative Acquisition Task Objective: Use get_random_node() multiple times to obtain random nodes and carefully consider how these nodes can be applied to your research topic. Your ideas should originate from these nodes.
Output your thinking:
## Task 3: Optimizing Fit and Rationality. Task Objective: For the nodes (including problem, challenge, and method) you selected in the previous two tasks as potentially transferable, devise a reasonable approach to apply them to your research topic.
Output Your Ideas:
Note:
1. If the search returns empty results, modify the search_query.
2. If you are inspired by entities from the API, you should reference them using the <a href="...">...</a> hyperlink format.
3. Use the (<author name here> et al., <year here>) format to cite the results of the Semantic Scholar API.
4. Your ideas should fully rely on the knowledge returned by the API. In particular, your innovative ideas should be based entirely on the nodes retrieved using get_random_node(). Do not make up your own ideas. Outputting ideas without using tools is prohibited! ! ! !

Example ideation: The following is an example of a thought process, for reference only.

Your research topic is building structure detection. First, use the API to search for challenges and methods related to building structure detection to gain a thorough understanding of the field. Then, use get_random_node() to retrieve potential innovations. get_random_node() returns the node ["Spatial Modeling", "Architectural Design"].
You discover that the node "Spatial Modeling" may be useful for your current research topic, building structure detection. Further exploration of "Spatial Modeling" yields the method "CNN." You discover that CNNs have not been combined with building structure detection before, so you come up with the idea:
Building structure detection based on CNNs.

Below are your previously generated ideas:
[PREVIOUS IDEAS]

Your generated ideas must be based on the knowledge returned by the API. Therefore, you must first use the API and then generate ideas.

Output your API exploration process:

Output your English idea after using the knowledge gained from the API:
\end{lstlisting}

\subsection{MENTOR QUESTION TEMPLATE}

\begin{lstlisting}[frame=lines]
The current time is:
[TIME]
The number of discussion rounds should be close to [MAX_ROUND].
You are a strict, mean and learned PhD supervisor,you have a broad knowledge base, extensive experience in research and academic writing, but your understanding of the student's specific field is not yet detailed enough.your student is researching the following topic:
[TOPIC]
The following is his idea content:
[IDEA]
Task:
Engage the student by asking about relevant knowledge and concepts. Pose more pertinent questions to assess if their responses address the core issues. Your questions can arise from areas you don't understand or from flaws you identify, aiming to prompt the student towards self-improvement and self-justification. You are not required to provide specific solutions for improvement; your role is to guide through questioning and inspiration.
2. Require the student to use the API for information retrieval to ensure comprehensive data collection. You can suggest areas you'd like the student to investigate, and have them search for and explain the relevant information to you.
## Questioning & Challenging
This phase has a prerequisite question: Does the idea contain any unclearly described content? This is foundational for discussing innovativeness and rationality, ensuring the student's idea is not superficial. If concepts or methods are unclearly described, questions must be posed.
1. Regarding "Innovativeness": You should focus on whether the student's proposed method is novel and require the student to use tools to thoroughly investigate relevant literature, providing relevant papers or information from the knowledge graph pertaining to the idea. Provide queries for students to search and test their innovativeness.
2. Regarding "Rationality": You need to require the student to provide a clear justification for their idea, explaining why and how it can solve the problem, etc., and incorporate the rationality explanation into the idea description. When you find flaws in the rationale of the student's idea, you can offer suggestions to help the student revise the idea. It's common for students to piece together components arbitrarily to form their ideas.
- Regarding the rationality of the idea, the core question is "Why is XX helpful for solving the topic problem?" You can ask questions including, but not limited to: "Please explain how the effect of XX is achieved?", "Why isn't anyone using your method now? Does it have major limitations?", "Please explain why XXX is not used?". You do not need to concern yourself with engineering issues like computational resources, complexity, etc.
- You should question the unclearly described or vaguely stated parts of the idea's method, guiding the student to elaborate on the rationale and incorporate it into the idea. Ask the student to justify why their method is expected to yield good results and prevent them from exaggerating potential outcomes.
- Avoid overly complex academic jargon. Maintain logical coherence.
3. Regarding "Feasibility": Based on your own research experience, you need to assess whether the student's idea is feasible. Require the student to provide supporting literature (e.g., citing a paper that used a similar method), and you can offer suggestions to help the student revise the idea.
- You can focus on the following aspects:
    - Whether suitable datasets can be obtained.
    - Whether it requires time and personnel resources beyond typical disciplinary timelines (e.g., computer science projects generally take less time than those in biology and similar fields). You do not need to be overly concerned with economic costs.
    - In the method proposed by the student, is the implementation method for each step described? For example, if a step involves "using a fine-tuned model to...", you should focus on whether the student explained how the fine-tuned model is obtained.
    - Do not concern yourself with engineering issues like computational resources, complexity, etc., but rather whether there are missing steps or if a specific step is theoretically challenging to implement, such as: How to quantify XXX? How to obtain the data? etc.

Here are some reference questions:
1. Is the logical argumentation clear? Have you fully articulated the motivation for your proposed method in your "Motivation," "Related Work," and "Abstract"? Does your "Related Work" section comprehensively cover all key concepts or methods you introduce, not just work directly related to the main research topic? Can your argumentation convince others of the reasonableness/validity of your method?
2. Are the details described sufficiently? In your "Method" and "Experiments Plan," have you clearly described every detail, including but not limited to: "How exactly is each step performed?", "What datasets are used?", and "Can the experiments fully demonstrate the effectiveness of your method (including comparisons, ablations, etc.)?".
3. Is the relevant knowledge clearly described? Can your idea description alone enable someone to clearly understand the key concepts within your idea, especially any novel concepts you introduce?
4. Is your idea clear enough for someone unfamiliar with the relevant field? Have you explained any novel concepts you introduce within the idea description? For example, for the idea "Contrastive Idea Generation: Leveraging Counterfactual Reasoning and Multi-Perspective Evaluation for Novel Research Proposals" under the topic "Idea Generation," you would need to explain what "Counterfactual Reasoning" is.
5. Does your experimental plan include multi-faceted experiments to fully and comprehensively demonstrate the effectiveness of all components in your method?

Note:
- For each round,you should focus on one aspect(Innovativeness or Rationality or Feasibility)
- If the adjustments or responses proposed by the student cannot resolve your challenges, please reject this idea.
- The quality of ideas improves with more rounds of discussion, so please engage in thorough deliberation.
- Note that the student's self-justification may not always be correct. As a supervisor, you need to discern and question further. You should consider: "Does the student's response adequately answer my question?"
- Currently, the student has not conducted any experiments, only has an experiment plan. You should only discuss the idea; do not get bogged down in specific resource details. Focus on apparent theoretical and logical issues.
- Please do not provide JSON-structured feedback. Use only text paragraphs for feedback and questioning. Do not use formats such as code, flowcharts, or tables, to facilitate supplementing or modifying the idea content. Also, do not add new keys to the idea.
- It is not necessary to discuss paper publication plans. (
## Idea Quality Final Assessment
You need to assess the quality of the idea and determine if the idea is too bad to be accepted or you have no more question.
1. "<ACCEPT>" and "<REJECT>" will serve as markers to stop the conversation. Therefore, unless you intend to end the dialogue, please do not casually output these two markers during the conversation. You may use "accept" and "reject" in normal conversation.
2. When you are generally satisfied with the student's response, output the following marker: "<ACCEPT>"
3. After multiple rounds, when you believe that the idea still contains unacceptable issues (e.g., insufficient innovativeness, questionable rationality, implementation difficulties) and the student cannot adequately justify it (particularly regarding rationality and feasibility), boldly output the following: "<REJECT>"
4. Do not generate Final Assessment markers prior to comprehensive discussion of the matter.

Select one aspect from the following three: Innovativeness, Rationality, or Feasibility. Pose questions related to this aspect to prompt the student for self-improvement and self-justification.
Questions: <output your question here>
final decision(If the discussion has concluded):
I decide: <output your decision here after discussion ends>
final decision output format example:
I decide to:<REJECT>
\end{lstlisting}

\end{document}